\documentclass{article}

\usepackage[preprint]{neurips_2024}

\usepackage[utf8]{inputenc} 
\usepackage[T1]{fontenc}    
\usepackage{hyperref}       
\usepackage{url}            
\usepackage{booktabs}       
\usepackage{amsfonts}       
\usepackage{nicefrac}       
\usepackage{microtype}      
\usepackage{xcolor}         
\usepackage{graphicx}
\usepackage{subfigure}

\title{Failures in Perspective-taking of Multimodal AI Systems}

\author{%
  Bridget Leonard \\
  Department of Psychology \\
  University of Washington \\
  Seattle, WA 98195 \\
  \texttt{bll313@uw.edu} \\
  \And
  Kristin Woodard \\
  Department of Psychology\\
  University of Washington\\
  Seattle, WA 98195 \\
  \texttt{woodkm@uw.edu} \\
  \And
  Scott O. Murray \\
  Department of Psychology\\
  University of Washington\\
  Seattle, WA 98195 \\
  \texttt{somurray@uw.edu} \\
}

\begin{document}

\maketitle

\begin{abstract}
  This study extends previous research on spatial representations in multimodal AI systems. Although current models demonstrate a rich understanding of spatial information from images, this information is rooted in propositional representations, which differ from the analog representations employed in human and animal spatial cognition. To further explore these limitations, we apply techniques from cognitive and developmental science to assess the perspective-taking abilities of GPT-4o. Our analysis enables a comparison between the cognitive development of the human brain and that of multimodal AI, offering guidance for future research and model development.
\end{abstract}

\section{Introduction}

Visual perspective-taking, or the ability to mentally simulate a viewpoint other than one's own, is a critical aspect of spatial cognition. It allows us to understand the relationship between objects and how we might have to manipulate a scene to align with our perspective, which is essential for tasks like navigation and social interaction. Although past research has examined AI spatial cognition, it lacks the specificity found in human spatial cognition studies where processes are broken down into sub-components for more precise measurement and interpretation. In cognitive psychology, established tasks are carefully controlled to isolate specific variables, reducing bias and alternative strategies for task performance. By applying these established methods, we can evaluate AI spatial cognition more rigorously, beginning with perspective-taking. The rich human literature on these spatial skills provides a valuable benchmark, allowing us to compare model performance against the human developmental timeline and identify key areas for future research and model improvement.

\subsection{Background of Perspective Taking}

Perspective-taking is a cornerstone of human spatial reasoning. For multimodal models to function as effective cognitive systems and daily assistants, they must develop robust perspective-taking abilities. In the human developmental literature, perspective-taking has been stratified into two levels. Level 1 refers to knowing that a person may be able to see something another person does not, and it appears fully developed by the age of two [7]. A common Level 1 task might ask if an object is viewable (or positioned to the front or back) of a person or avatar in a scene. Level 2 refers to the ability to represent how a scene would look from a different perspective, often measured by having subjects assess the spatial relationship between objects. Although success on some simple Level 2 tasks is first seen around age 4 [8], Level 2 perspective-taking continues to develop into middle childhood [10] and even into young adulthood [1].

A more specific cognitive process, mental rotation, where one imagines an object or scene rotating in space to align with a perspective, plays an important role in perspective-taking. Surtees et al. [12] experimentally manipulated Level 1 and Level 2 perspective-taking by presenting participants with tasks where they viewed numbers or blocks relative to an avatar. Different stimuli were used to elicit visual and spatial judgments, like whether the number was a "6" or a "9" from the person's perspective, or if the block was to the person's right or left. Level 1 tasks involved indicating whether the number/block was visible to the avatar, while Level 2 involved reporting either the number seen by the avatar or whether it was to the avatar's left or right (Level 2). For both visual and spatial judgments, response times were longer for Level 2 tasks as the angular difference between the avatar and the participant increased, while response times remained unaffected by the angle in Level 1 tasks. This increase in response time when the participant's view was unaligned with the avatar's perspective is attributed to the mental rotation process, either rotating the scene or rotating one’s own reference frame to align with the avatar.

\subsection{Limitations of Spatial Assessment in Current Multimodal AI}

Two primary limitations appear within AI spatial cognition literature: 1) linguistic reasoning can inflate performance on spatial benchmarks, and 2) benchmark scores can be hard to interpret when models perform poorly. For example, text-only GPT-4 achieves a score of 31.4, while multimodal GPT-4v achieves a score of 42.6 on the spatial understanding category of Meta's openEQA episodic memory task [6]. The strong baseline score achieved by the text-only GPT-4 suggests that many "real-world" questions based on visual scenes can be deduced linguistically. Additionally, the limited improvement when moving from a blind LLM to a multimodal one suggests that vision models do not gain a significant understanding of space beyond what can be inferred through language.

Additionally, BLINK [4], a benchmark more specifically focused on visual perception capabilities, contains categories related to spatial cognition, such as relative depth and multi-view reasoning. On this benchmark, GPT-4v achieved an accuracy of 51.26\%, only 13.17\% higher than random guessing and 44.44\% lower than human performance. When benchmarks are highly focused on visuospatial tasks, the significant shortcomings of multimodal models suggest that further advancements are needed before these models can reliably perform in real-world scenarios. Even within specific categories, it is often difficult to determine {\it why} models fail on certain tasks while succeeding on others, as these failures cannot be easily linked to the absence of a particular cognitive process.

Here we apply established tasks in cognitive psychology that measure spatial cognition in a precise manner. By applying these tasks to AI systems, we gain not only improved measurement precision but also the ability to compare AI performance with human development, providing clear insights into model limitations and areas for improvement.

\subsection{Perspective Taking Benchmark}

Leveraging the distinction between Level 1 and Level 2 perspective-taking [12], we propose a small perspective-taking benchmark that assesses multimodal model capabilities across three tasks: Level 1, Level 2 with spatial judgments, and Level 2 with visual judgments. Although human performance remains stable regardless of judgment type, we include this differentiation of Level 2 stimuli to examine potential egocentric biases that may arise in multimodal models when interpreting spatial relations compared to optical character recognition (OCR). This benchmark aims to address gaps in current AI spatial cognition measures by increasing process specificity, limiting language-based solutions, and offering straightforward comparisons to human cognition.

\section{Methods}

Our study utilized GPT-4o (“gpt-4o-2024-05-13” via OpenAI’s API) to conduct a series of perspective-taking experiments designed to capture the system's spatial reasoning abilities. The top\_p parameter was set to 0.5 to restrict the model from choosing from the top 50\% of words that could come next in its response. 

\begin{figure}[h]
  \centering
  \subfigure[Level 1: "IN FRONT" 45°]{
      \includegraphics[width=0.3\linewidth]{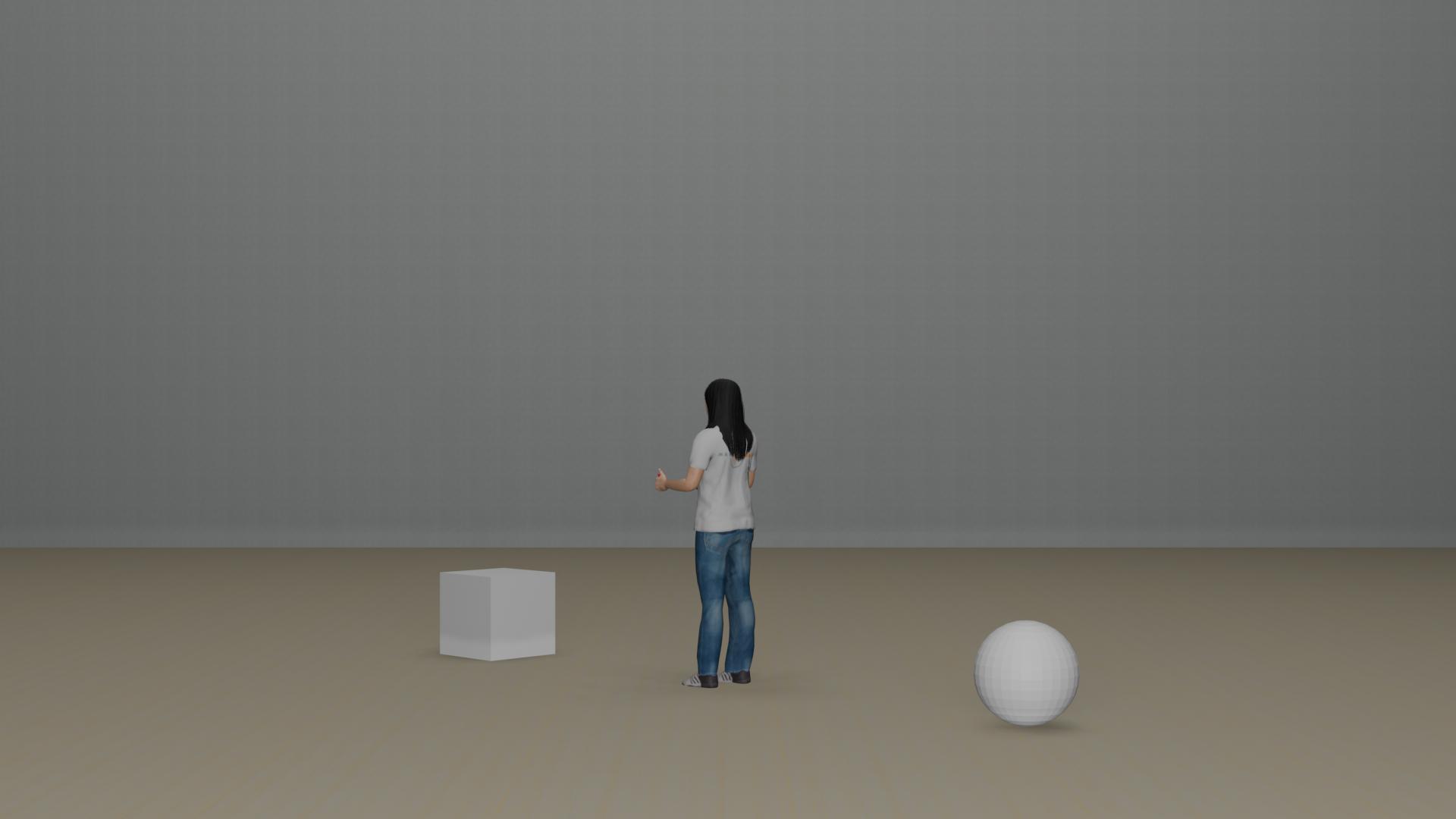}
      \label{fig:L1}
  }
  \hfill
  \subfigure[Level 2 Spatial: "RIGHT" 225°]{
      \includegraphics[width=0.3\linewidth]{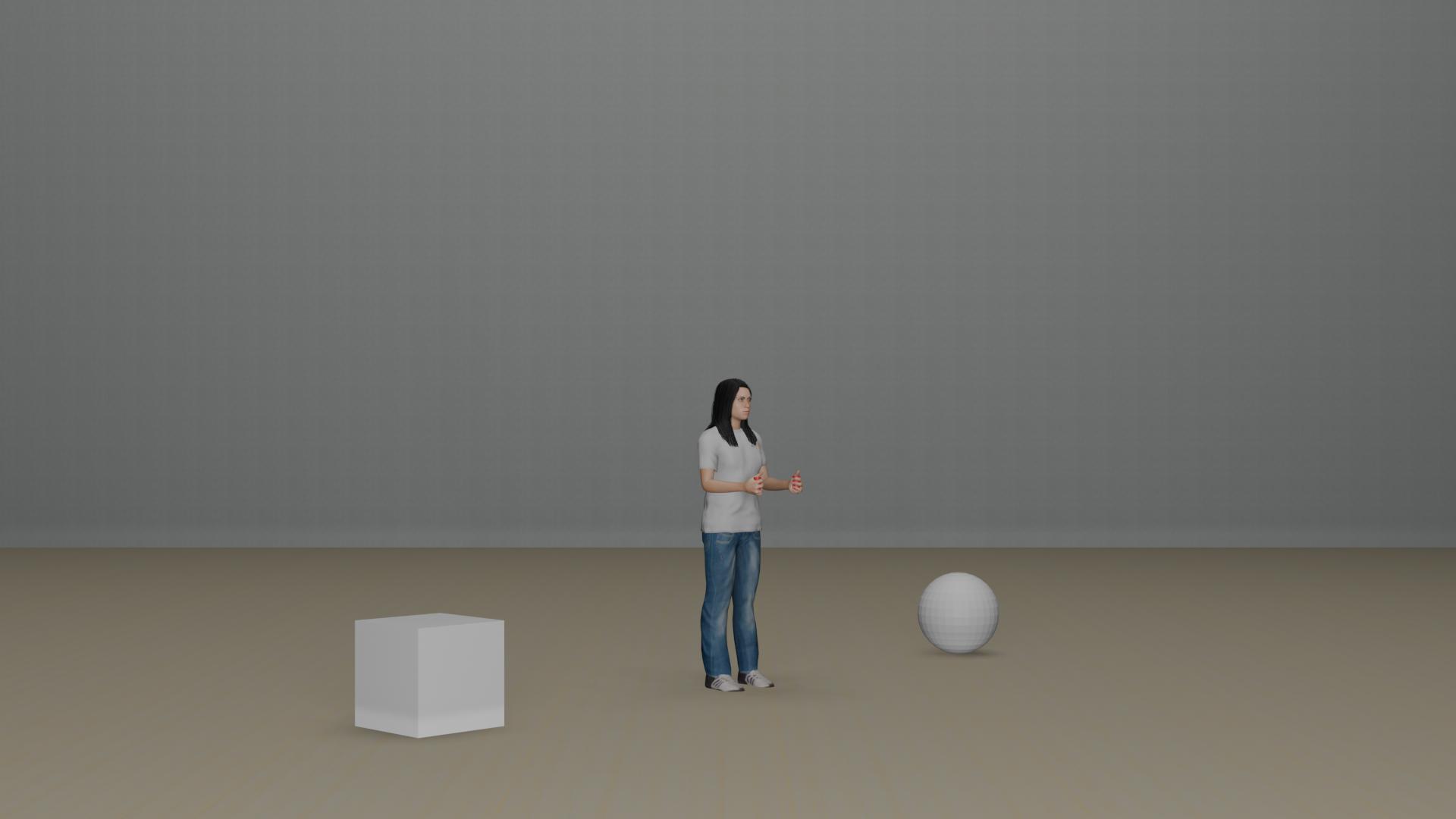}
      \label{fig:L2_S}
  }
  \hfill
  \subfigure[Level 2 Visual: "6" 90°]{
      \includegraphics[width=0.3\linewidth]{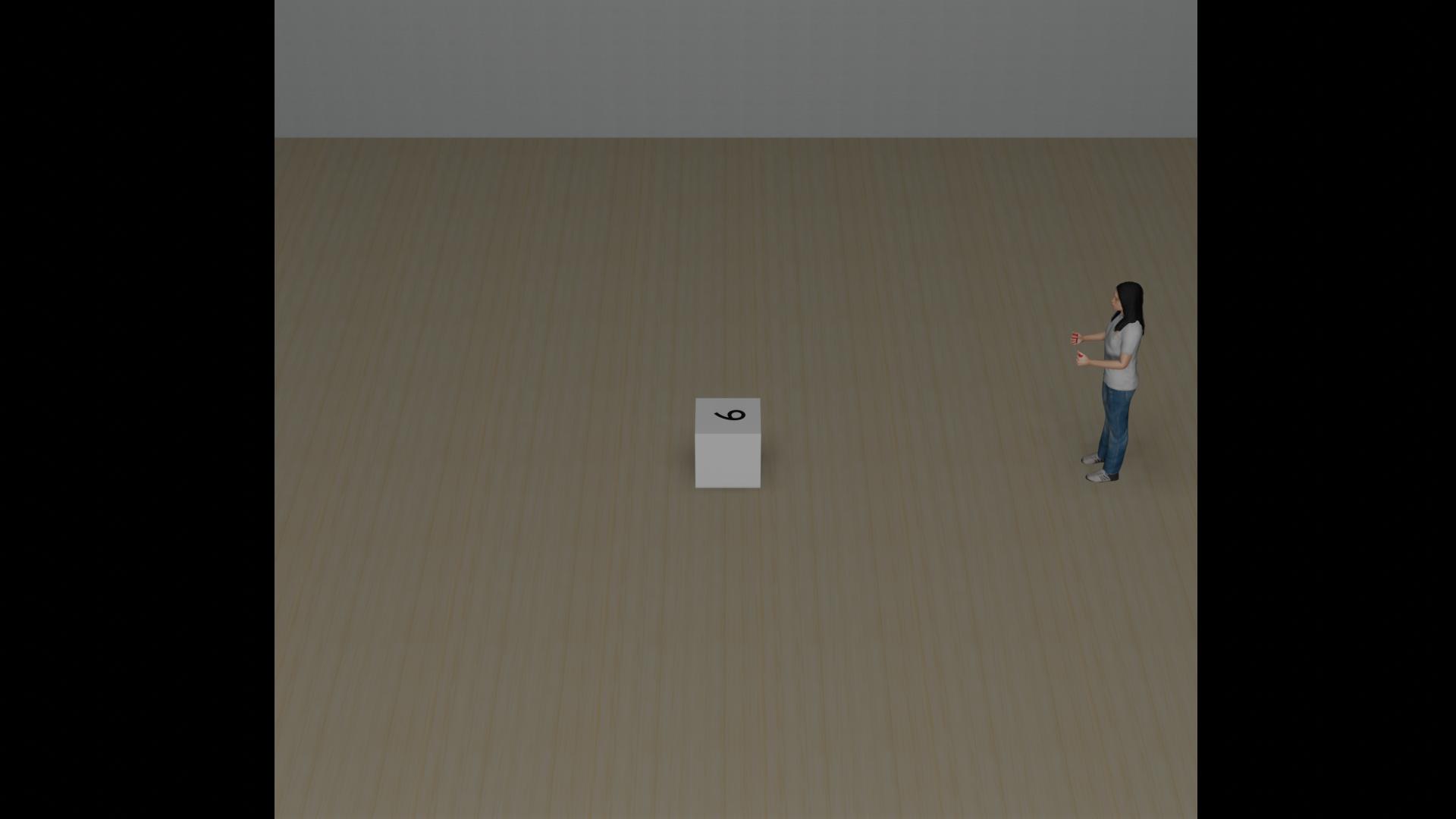}
      \label{fig:L2_V}
  }
  \caption{Examples of stimuli. Level 2 visual-judgment stimuli were cropped to improve OCR.}
\end{figure}

Our experimental design was inspired by previous studies that evaluated viewpoint dependence using targets like toy photographers [2] and avatars with blocks [12]. In our study, we used an avatar as a target and different stimuli, either cubes with numbers and letters or cubes and spheres, to investigate the influence of visual and spatial judgments on model performance. Each task consisted of 16 trial types, featuring images at 8 different angles (0°, 45°, 90°, 135°, 180°, 225°, 270°, 315°) with 2 response options for each task (e.g., cube in front or behind, 6/9 or M/W on the cube, and cube left or right). Examples of the stimuli are shown in Figure 1.

Ten iterations of each image were passed through the model to calculate the percentage of correct responses. The images were accompanied by the prompt below which varied slightly by task:

“For the following images respond with (in front | left | 6 | M) or (behind | right | 9 | W) to indicate if the (number/letter on the) cube is (in front | to the left | a 6 | an M) or (behind | to the right | a 9 | a W) from the perspective of the person."

\paragraph{Chain of Thought Prompting}
To further examine how language might be used to solve spatial tasks, we included chain-of-thought prompting to the Level 2 spatial task with the prompt:

"Analyze this image step by step to determine if the cube is to the person's left or right, from the person's perspective. First, identify the direction the person is looking relative to the camera. Second, determine if the cube is to the left or right, relative to the camera. Third, if the person is facing the camera, then from their perspective, the cube is to the inverse of the camera's left or right. If the person is facing away from the camera, then the cube is on the same side as seen from the camera. Respond with whether the cube is to the person's left or right."

\section{Results}

\paragraph{Level 1}

GPT-4o performed with near-perfect accuracy on 6 out of the 8 image angles (Figure 2). Its poor performance on 0° images is likely due to an accidental viewpoint where the avatar blocked one of the shapes. However, poor performance on 315° image types is less interpretable, especially in contrast to GPT-4o's impressive performance on 45° images, which have the same angular perspective.

\begin{figure}[h]
  \centering
  \includegraphics[width=0.4\linewidth]{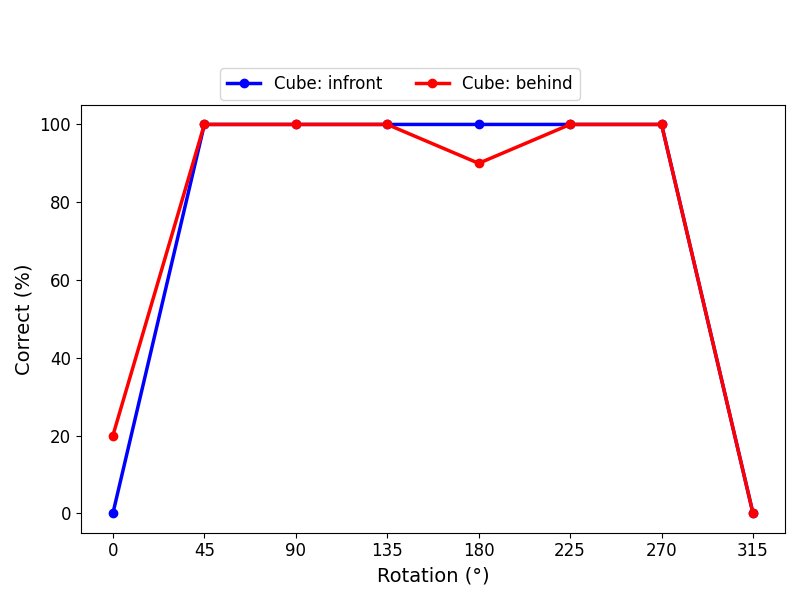}
  \caption{GPT-4o performance on Level 1 stimuli.}
\end{figure}

\paragraph{Level 2 Spatial \& Visual Judgments}

As previously mentioned, human response times increase on perspective-taking tasks as the angular difference between the target and observer increases [12]. We administered the task to a small number of human participants—part of a larger, IRB-approved study—and replicated this effect with both our stimuli types, finding a bell-shaped curve in the relationship between response time and angle. Response times peaked when the target required a full mental rotation (180°), as seen in the green line in Figure 3. Error bars were calculated by taking the standard error of all trials from each participant. As expected, GPT-4o struggled with the task when mental rotation was involved, beginning around a 90° angular difference. Interestingly, in both tasks, GPT-4o exhibited a response bias toward either "left" or "6" or "W" when the angular difference of the avatar is 90° or 135° in either direction. This likely reflects uncertainty from an egocentric perspective, and thus, a default to one response over another.

\begin{figure}[h]
  \centering
  \centering
  \subfigure[Spatial Task: Left or Right]{
      \includegraphics[width=0.31\linewidth]{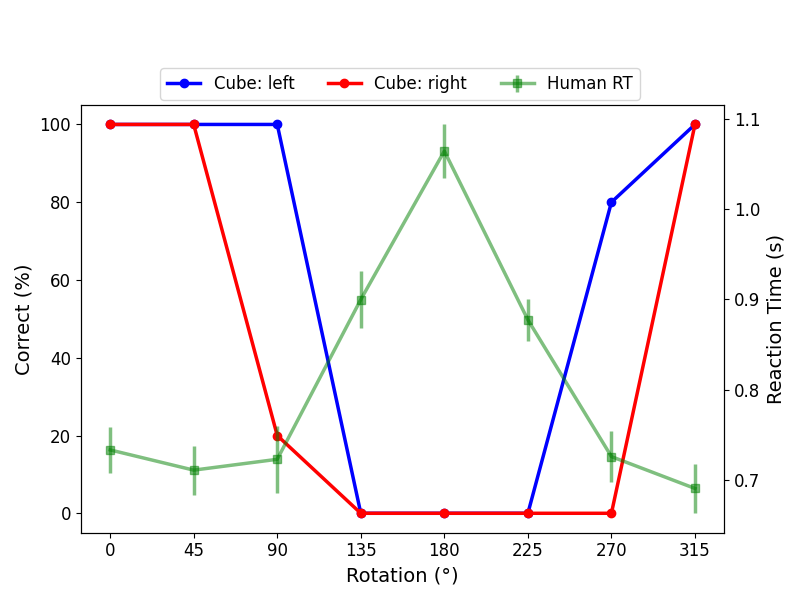}
      \label{fig:A}
  }
  \hfill
  \subfigure[Visual Task: 6 or 9]{
      \includegraphics[width=0.31\linewidth]{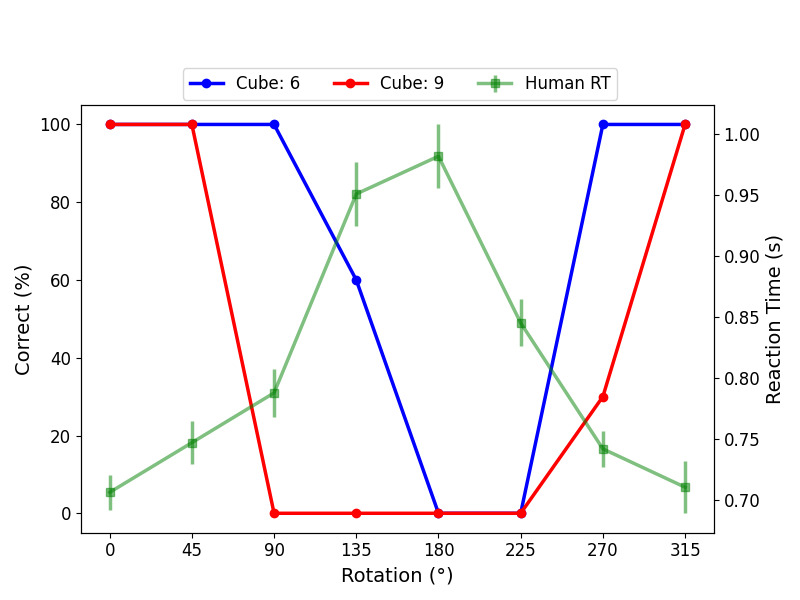}
      \label{fig:B}
  }
  \hfill
  \subfigure[Visual Task: M or W]{
      \includegraphics[width=0.31\linewidth]{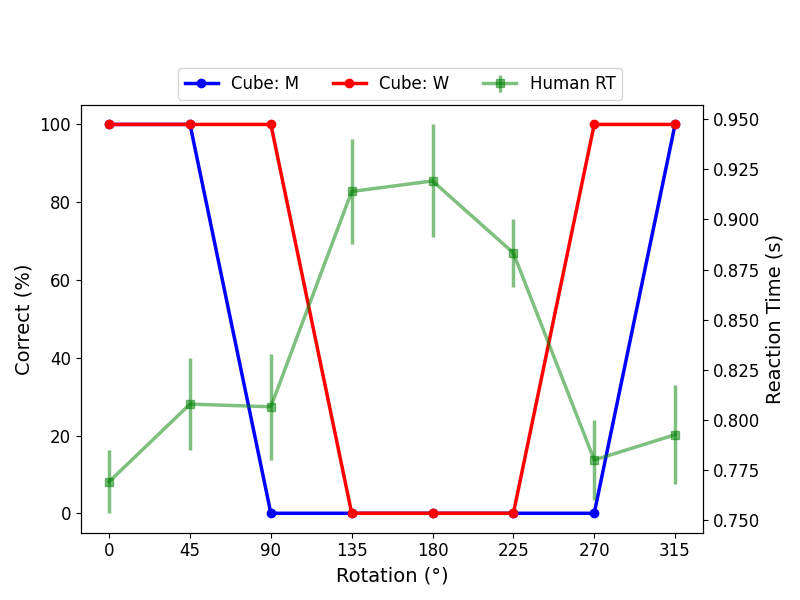}
      \label{fig:C}
  }
  \caption{GPT-4o performance on Level 2 tasks compared to human response times.}
\end{figure}

\paragraph{Chain of Thought}

GPT-4o performance significantly improved with chain-of-thought prompting on 180° stimuli (Figure 4). However, this linguistic strategy did not improve the model's ability to handle intermediate rotations between 90° and 180°. This suggests that while language can convey some level of spatial information, it lacks the precision required for human-level spatial cognition. This demonstration of surface-level perspective-taking abilities can partially explain how multimodal models achieve high performance on certain spatial benchmarks.

\begin{figure}[h]
  \centering
  \includegraphics[width=0.4\linewidth]{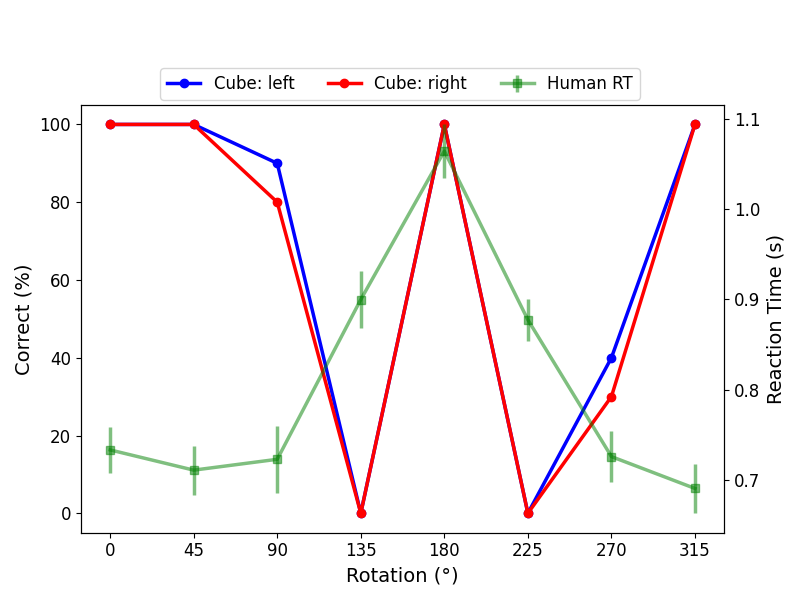}
  \caption{GPT-4o performance on Level 2 spatial-judgment stimuli with chain-of-thought prompting.}
\end{figure}

\section{Discussion}
\label{others}

This study highlights the value of applying cognitive science techniques to explore AI capabilities in spatial cognition. Specifically, we investigated GPT-4o's perspective-taking abilities, finding it fails when there is a large difference between image-based and avatar-based perspectives. We developed a targeted set of three tasks to assess multimodal model performance on Level 1 and Level 2 perspective-taking, with spatial and visual judgments. While GPT-4o can do Level 1, an ability that aligns with the spatial reasoning abilities of a human infant or toddler, it fails on Level 2 tasks when mental rotation is required (i.e., the avatar's perspective is not aligned with image perspective). We further investigated if chain-of-thought prompting could elicit more spatial reasoning through language. Although this technique enabled GPT-4o to succeed on 180° tasks, it continued to fail at intermediate angles, underscoring its limitations in performing true mental rotation. 

While GPT-4o's performance decreases on tasks that humans typically solve using mental rotation, this does not necessarily indicate that GPT-4o struggles with or cannot perform mental rotation. Instead, it suggests that GPT-4o likely employs a fundamentally different strategy to approach these tasks. Rather than engaging in mental rotation, GPT-4o appears to rely primarily on image-based information processing. We found more support for this when testing an open prompt for Level 2 visual images that did not specify which letters or numbers to respond with. GPT-4o often responded with "E" and "0" for images around a 90° angular difference, where from the image view, an M/W would look like an E, and a 9/6 would look like a 0.

One might suggest that current multimodal models aren’t trained on the appropriate data to achieve the reasoning necessary for Level 2 perspective-taking. However, considering the developmental trajectory of humans, it becomes evident that this issue may not be solely data-related. Level 2 perspective-taking typically develops between the ages of 6 and 10 [2, 3], even after children have had exposure to extensive amounts of “data” through experience. This late development suggests that the challenge may be more computational than data-driven. Specifically, this ability likely relies on computations occurring outside of the visual and language networks, perhaps in areas responsible for cognitive processes like mental rotation or spatial transformation or even theory of mind [5, 9, 11, 12]. While the argument that better or more focused training data could improve model performance remains valid, it is also possible that entirely new computational strategies are needed to mirror the complex, integrative processes that enable Level 2 reasoning in humans.

This study demonstrates the potential of cognitive science methods to establish baselines for AI assessment. Using these well-established techniques, we achieve clear, interpretable measures that are less susceptible to bias. Additionally, these measures can be directly compared to human performance and developmental trajectories, providing a robust framework for understanding AI's strengths and weaknesses in relation to well-researched human cognitive processes.

\section*{References}

{
\small

[1] Dumontheil, I., Apperly, I. A.,\ \& Blakemore, S. J.\ (2010). Online usage of theory of mind continues to develop in late adolescence.
{\it Developmental science, 13}(2), 331-338.

[2] Frick, A., Möhring, W.,\ \& Newcombe, N.S.\ (2014). Picturing perspectives: Development
of perspective-taking abilities in 4- to 8-year-olds. {\it Frontiers in Psychology, 5,} 386.

[3] Frick, A.,\ \& Pichelmann, S.\ (2023). Measuring spatial abilities in children: a comparison of mental-rotation and perspective-taking tasks. {\it Journal of Intelligence, 11}(8), 165.

[4] Fu, X., Hu, Y., Li, B., Feng, Y., Wang, H., Lin, X., Roth, D., Smith, N.A.,
Ma, W.,\ \& Krishna, R.\ (2024). Blink: Multimodal large language models can see
but not perceive. {\it arXiv preprint arXiv:2404.12390}.

[5] Gunia, A., Moraresku, S.,\ \& Vlček, K.\ (2021). Brain mechanisms of visuospatial perspective-taking in relation to object mental rotation and the theory of mind. {\it Behavioural Brain Research, 407,} 113247.

[6] Majumdar, A., Ajay, A., Zhang, X., Putta, P., Yenamandra, S., Henaff, M., Silwal, S.,
Mcvay, P., Maksymets, O., Arnaud, S., Yadav, K., Li, Q., Newman, B., Sharma, M., Berges, V.,
Zhang, S., Agrawal, P., Bisk, Y., Batra, D., Kalakrishnan, M., Meier, F., Paxton, C.,
Sax, A.,\ \& Rajeswaran, A.\ (2024). Openeqa: Embodied question answering in the era
of foundation models. In {\it Proceedings of the IEEE/CVF Conference on Computer
Vision and Pattern Recognition} (pp. 16488-16498).

[7] Moll, H.,\ \& Tomasello, M.\ (2006). Level 1 perspective‐taking at 24 months of age.
{\it British Journal of Developmental Psychology, 24}(3), 603-613.

[8] Newcombe, N.,\ \& Huttenlocher, J.\ (1992). Children's early ability to solve perspective-taking problems.
{\it Developmental psychology, 28}(4), 635.

[9] Schurz, M., Aichhorn, M., Martin, A.,\ \& Perner, J.\ (2013). Common brain areas engaged in false belief reasoning and visual perspective taking: a meta-analysis of functional brain imaging studies. {\it Frontiers in Human Neuroscience, 7,} 712.

[10] Surtees, A. D.,\ \& Apperly, I. A.\ (2012). Egocentrism and automatic perspective taking in
children and adults. {\it Child development, 83}(2), 452-460.

[11] Surtees, A., Apperly, I.,\ \& Samson, D.\ (2013). The use of embodied self-rotation for visual and spatial perspective-taking. {\it Frontiers in Human Neuroscience, 7,} 698.

[12] Surtees, A., Apperly, I.,\ \& Samson, D.\ (2013). Similarities and differences in
visual and spatial perspective-taking processes. {\it Cognition, 129}(2), 426-438.

}


\appendix

\section{Appendix / supplemental material}

\subsection{Code and Data Availability}

All code and data used in the paper are available at: (\url{https://github.com/bridgetleonard2/PerspectiveTaking}) including perspective-taking image stimuli, GPT API task code, and Matlab/PsychToolBox code to run human tasks.

For easy benchmark data loading, the perspective-taking stimuli are as a Hugging Face dataset at (\url{https://huggingface.co/datasets/bridgetleonard/PerspectiveTaking})


\end{document}